\pdfoutput=1

\documentclass[11pt]{article}

\usepackage{ACL2023}

\usepackage{times}
\usepackage{latexsym}

\usepackage[T1]{fontenc}

\usepackage[utf8]{inputenc}

\usepackage{microtype}

\usepackage{inconsolata}

\usepackage{multirow}
\usepackage{multicol}
\usepackage{amssymb}
\usepackage{subcaption}
\usepackage{bm}
\usepackage{cite}
\usepackage{booktabs}
\usepackage{graphics}
\usepackage{amsmath,epsfig}

\usepackage{bbding}
\usepackage{amssymb}
\usepackage{pifont}

\usepackage{xcolor}

\usepackage{algorithm}
\usepackage[noend]{algorithmic}
\usepackage{eqparbox}

\definecolor{visualorange}{HTML}{F4B183}
\definecolor{textgreen}{HTML}{70AD47}
\definecolor{applegreen}{rgb}{0.55, 0.71, 0.0}

\usepackage{makecell}
\usepackage{fancybox}
\usepackage[many]{tcolorbox}

\definecolor{mypattern}{RGB}{222,235,247}
\newtcbox{\pattern}{on line,colback=mypattern,colframe=white,size=fbox,arc=3pt, box align=base, before upper=\strut, top=-2pt, bottom=-2pt, boxrule=0pt}

\definecolor{myentity}{RGB}{255,242,204}
\newtcbox{\entity}{on line,colback=myentity,colframe=white,size=fbox,arc=3pt, box align=base, before upper=\strut, top=-2pt, bottom=-2pt, boxrule=0pt}

\usepackage{rotating}

\title{From Alignment to Entailment: \\ A Unified Textual Entailment Framework for Entity Alignment}

\author{Yu Zhao\textsuperscript{1}\;\;\;Yike Wu\textsuperscript{2} \;\;\;Xiangrui Cai\textsuperscript{1}\;\;\;Ying Zhang\textsuperscript{1}\thanks{\; Corresponding author.}  \\ {\bf Haiwei Zhang\textsuperscript{3}\;\;\;Xiaojie Yuan\textsuperscript{1}} \\
\textsuperscript{1} College of Computer Science, TKLNDST, Nankai University, Tianjin, China \; \\
\textsuperscript{2} School of Journalism and Communication, CMRC, Nankai University, Tianjin, China \; \\
\textsuperscript{3} College of Cyber Science, TKLNDST, Nankai University, Tianjin, China \; \\
{\tt zhaoyu@dbis.nankai.edu.cn} \\ {\tt \{wuyike,caixr,yingzhang,zhhaiwei,yuanxj\}@nankai.edu.cn}}

\begin{document}
\maketitle
\begin{abstract}
Entity Alignment (EA) aims to find the equivalent entities between two Knowledge Graphs (KGs).
Existing methods usually encode the triples of entities as embeddings and learn to align the embeddings, which prevents the direct interaction between the original information of the cross-KG entities.
Moreover, they encode the relational triples and attribute triples of an entity in heterogeneous embedding spaces, which prevents them from helping each other.
In this paper, we transform both triples into unified textual sequences, and model the EA task as a bi-directional textual entailment task between the sequences of cross-KG entities. 
Specifically, we feed the sequences of two entities simultaneously into a pre-trained language model (PLM) and propose two kinds of PLM-based entity aligners that model the entailment probability between sequences as the similarity between entities.
Our approach captures the unified correlation pattern of two kinds of information between entities, and explicitly models the fine-grained interaction between original entity information. 
The experiments on five cross-lingual EA datasets show that our approach outperforms the state-of-the-art EA methods and enables the mutual enhancement of the heterogeneous information.
Codes are available at \href{https://github.com/OreOZhao/TEA}{https://github.com/OreOZhao/TEA}.
\end{abstract}

\section{Introduction}

\begin{figure}[t]
     \centering
     \begin{subfigure}[b]{0.48\textwidth}
         \centering
         \includegraphics[width=\textwidth]{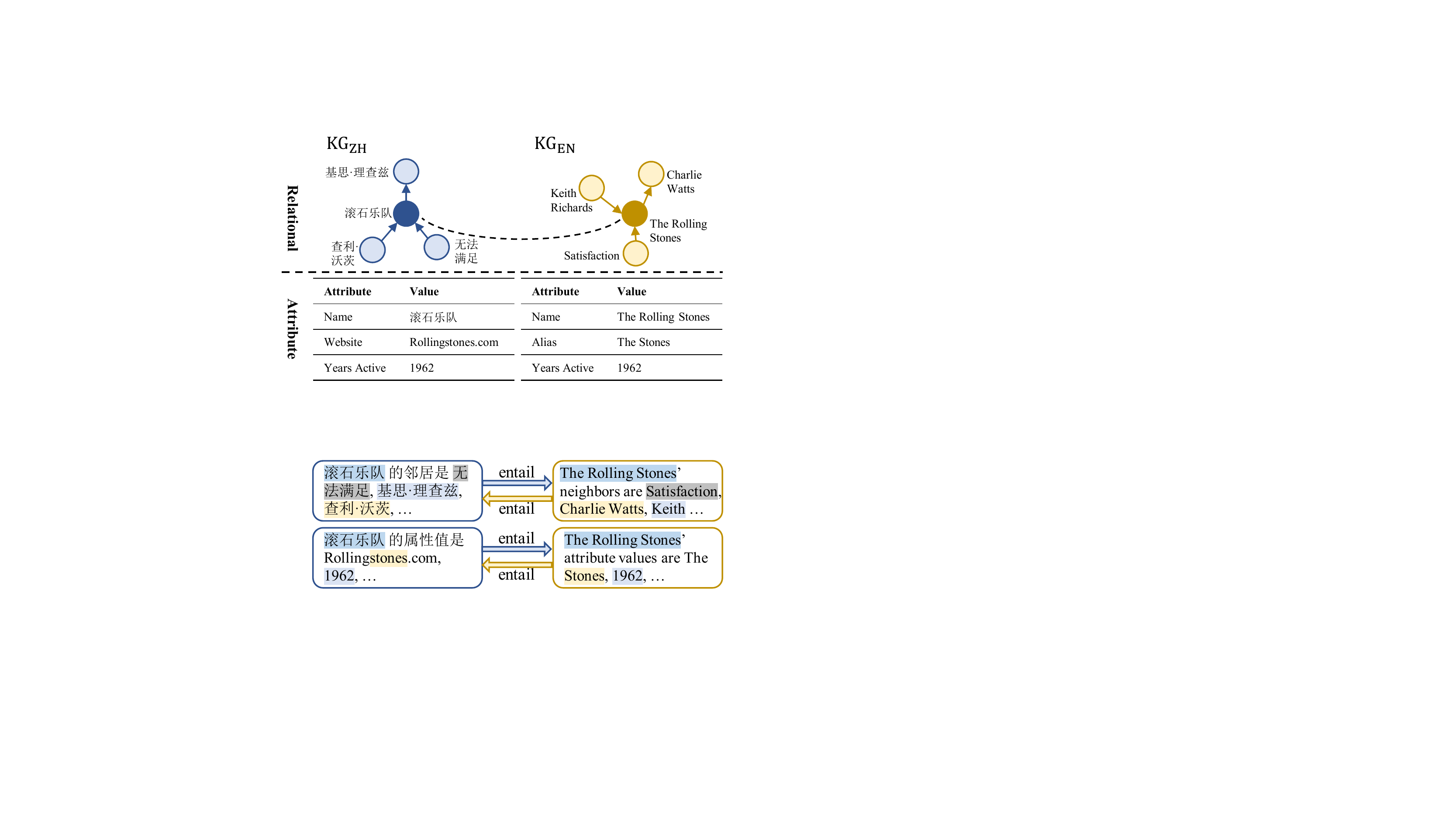}
         \caption{An example of heterogeneous relational and attribute information of entity "\textit{The Rolling Stones}" in ZH-EN KGs.}
         \label{fig:intro_hetero}
     \end{subfigure}
    \vspace*{\fill}
     \begin{subfigure}[b]{0.48\textwidth}
         \centering
         \includegraphics[width=\textwidth]{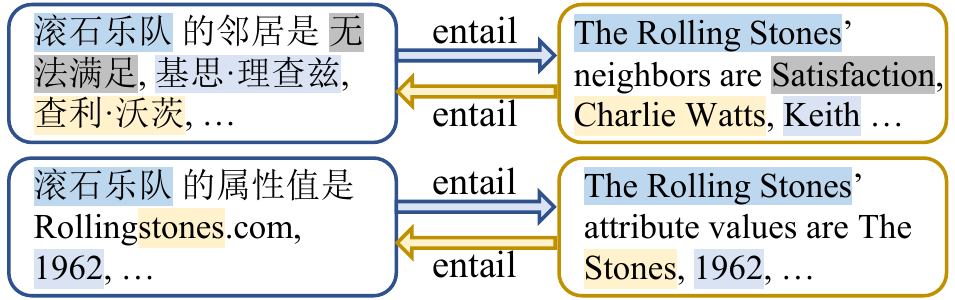}
         \caption{Our bi-directional entailment modeling of cross-KG entity sequences, where the sub-sequences with the same color shading share the same semantics.}
         \label{fig:intro_entail}
     \end{subfigure}
     \caption{(a) displays an example of relational and attribute information of entities. (b) displays our bi-directional entailment modeling for EA.}
\label{fig:intro}
\end{figure}
Knowledge Graphs (KGs) organize and store the facts in the real world to an effective structure, and have been applied to many knowledge-driven tasks, such as question answering \citep{lan2021kbqa}, recommender systems \citep{wang2022convrec}, and information extraction \citep{zhang2021kbie}. 
Since the KGs are often from various domains, Entity Alignment (EA) provides fundamental techniques to find the equivalent entities in two KGs, which would complement the knowledge coverage of KGs.

Existing EA methods usually consist of two modules: 
(1) embedding module encodes entity information to entity embeddings, (2) alignment module guides the embeddings of the aligned entities to be similar \citep{sun2020benchmarking}. 
Moreover, they usually incorporate two kinds of heterogeneous triples as shown in Figure \ref{fig:intro_hetero}: 
(1) relational triples $(h, r, t)$, represents the relation $r$ between head entity $h$ and tail entity $r$,
(2) attribute triples $(e, a, v)$, represents the attribute value $v$ of the attribute $a$ of entity $e$.

Despite the progress of existing EA methods \citep{liu2020attrgnn,tang2021bertint,zhong2022sdea}, they are limited by the embedding-based architecture in two folds:
\textbf{(1) Lack of direct interaction between KGs. }
Existing methods usually treat EA as a representation learning task.
During the encoding process, the origin triples of entities are compressed to a continuous vector, which prevents them from directly interacting with each other. 
However, the origin information contains rich semantics information.
Take the entity "\textit{The Rolling Stones}" in Figure \ref{fig:intro_hetero} as an example, the attribute value "\textit{Rollingstones.com}" and "\textit{1962}" of the Chinese KG are highly compatible with the value "\textit{The Rolling Stones}" and "\textit{1962}" in the English KG.
The correlation between the values can directly indicate the alignment of two entities.

\textbf{(2) Heterogeneous embedding spaces. }
Existing methods usually encode the relational triples and attribute triples in different embedding spaces due to the heterogeneity of structures and literals.
This way, the alignment of relational information and of attribute information are separated and could not help each other.
However, they may share the same correlation pattern.
For example, the entity "\textit{The Rolling Stone}" in Chinese and English KGs in Figure \ref{fig:intro} have common neighbors (translated) and common attribute values, which could both indicate the equivalence of entities.
Capturing the correlation pattern in a unified model would enable mutual enhancement between the two information.

Inspired by recent progress of pre-trained language models (PLMs) \citep{brown2020gpt3,gao2021lmbff,sun2022nspbert}, 
we transform both two kinds of triples into textual sequences, and propose a unified \textbf{T}extual \textbf{E}ntailment framework for entity \textbf{A}lignment \textbf{TEA}.
We model the EA task as a bi-directional textual entailment task between the sequences of cross-KG entities as shown in Figure \ref{fig:intro_entail} to explicitly capture the fine-grained interaction between entity information.
Specifically, we combine two sequences of entities in one sequence with cloze-style templates and feed the combined sequence into a PLM.
We further propose two aligners to model the entailment probability as the pre-training tasks of PLM, i.e. Next Sentence Prediction (NSP) and Masked Language Modeling (MLM).
The NSP-Aligner predicts the probability of whether one entity \textit{is next sentence} of the other, while the MLM-Aligner fills in the blanks between entity sequences with mapped label words "\textit{Yes}" or "\textit{No}".
The positive entailment probability is seen as entity similarity and is used for ranking the candidate entities.
The experiments on five cross-lingual EA datasets show that TEA outperforms the state-of-the-art methods and enables the mutual enhancement of heterogeneous information.

Overall, the contributions of this paper can be summarized as follows:
\begin{itemize}
    \item We unify the modeling of the relational triples and attribute triples in EA by transforming both into textual sequences and capturing their common correlation pattern.
    \item  To the best of our knowledge, we are the first to transform EA to a bi-directional textual entailment task of relational and attribute information. The proposed PLM-based aligners capture the fine-grained interaction between cross-KG entities.
    \item Experiments on five cross-lingual EA datasets demonstrate that our approach outperforms baselines and enables the mutual enhancement of heterogeneous information.
\end{itemize}

\section{Related Work}
\subsection{Entity Alignment} \label{sec: EA_related_work}

Existing EA methods usually follow an embedding-alignment architecture \citep{sun2020benchmarking}, where the entity encoder learns from the relational and attribute triples with various networks, then the alignment module guides the embeddings of the aligned entities to be similar. 

There are two mainstreams of methods: TransE \citep{bordes2013TransE} based methods \citep{chen2017MTransE,sun2017jape,zhu2017IPTransE,sun2018BootEA,guo2019RSN} for KG representation with simple implementation, and GCN \citep{welling2016originGCN} based methods \citep{chen2017MTransE,sun2017jape,zhu2017IPTransE,sun2018BootEA,guo2019RSN} for modeling graph structures.
However, the rich semantics in the origin information of cross-KG entities lack interaction through the encoding process.
Our work focuses on modeling the interaction between the origin information of cross-KG entities.

For methods incorporating attribute information with relational information, they usually encode them in heterogeneous representation spaces with hybrid encoders.
For example, GNNs \citep{sun2019transedge,liu2020attrgnn} and RNNs \citep{guo2019RSN,zhong2022sdea} are used for encoding relational triples to model the structures of entities, while Skip-gram \citep{sun2017jape}, N-hot \citep{wang2018GCNAlign,yang2019hman} and BERT \citep{liu2020attrgnn,zhong2022sdea} for attribute triples for capturing literal semantics. 
Some methods further aggregate the heterogeneous embeddings in separate sub-graphs \citep{wang2018GCNAlign,yang2019hman,liu2020attrgnn,tang2021bertint}.
However, the heterogeneous embedding spaces hinder the EA process.
Our work focuses on the unified modeling of relational and attribute information.

There have been other advancements in EA, focusing on unsupervised or self-supervised EA \citep{mao2021SEU,liu2022selfkg}, incorporation of entity images \citep{liu2021EVA,lin2022MCLEA}, EA with dangling cases \citep{sun2021dangling}, which motivates our future work.

\subsection{PLMs in KGs}
With the prosperity of PLMs like BERT \citep{devlin2019bert}, fine-tuning the PLM in downstream tasks has shown great potential in KGs.
In EA, several methods have explored PLMs in learning entity embeddings \citep{yang2019hman,tang2021bertint,zhong2022sdea}.
However, they share the same drawbacks with methods in Section \ref{sec: EA_related_work}, and some methods \citep{yang2019hman,tang2021bertint} require extra natural language sequences such as entity descriptions which are not always available.

Recent studies \citep{brown2020gpt3,gao2021lmbff,sun2022nspbert} show that given a natural-language prompt, the PLM could achieve remarkable improvements by simulating the pre-training tasks of PLM, i.e. NSP and MLM.
The prompt-based fine-tuning paradigm has been applied in many tasks in KGs, such as Named Entity Recognition \citep{huang2022promptner}, Entity Linking \citep{sun2022nspbert}, Entity Typing \citep{ding2021promptfget}.
However, there is no prompt-learning study for entity-pair tasks such as EA.
Our work focuses on constructing entity-pair sequences with prompts, and transforming the EA task to the NSP-style or MLM-style textual entailment task.
The entailment probability is seen as entity similarity.

\begin{figure*}[!t]
    \centering
    \includegraphics[width=1\textwidth]{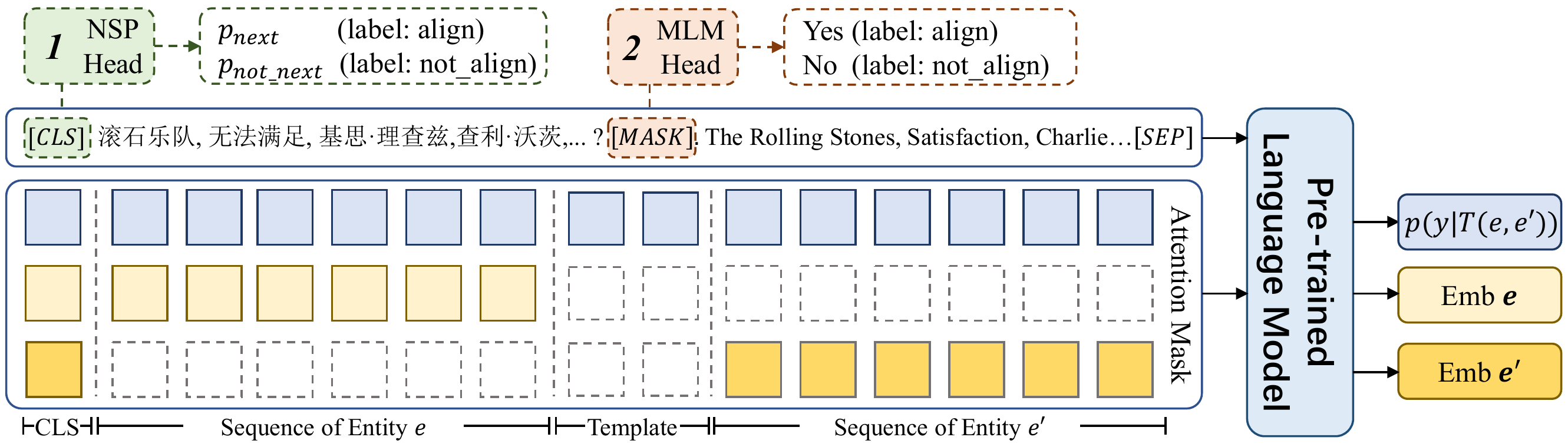}
    \caption{The architecture of TEA, textual entailment framework for entity alignment. 
    The input of PLM is the entity-pair sequence of $(e,e')$ and the attention mask we design for tuning PLM with both entailment and embedding-alignment objectives.
    The output of PLM is the probability of entailment $p(y|T(e,e'))$ and the embeddings of two entities $\textbf{e}$ and $\textbf{e}'$. The probability is from either NSP Head of NSP-Aligner or MLM Head of MLM-Aligner.
    }
    \label{fig: model_overview}
\end{figure*}

\section{Methodology}
\subsection{Preliminaries}
\textbf{Knowledge Graph. }
A knowledge graph (KG) could be defined as $\mathcal{G} = \{\mathcal{E}, \mathcal{R}, \mathcal{A}, \mathcal{V}, \mathcal{T}^r, \mathcal{T}^a\}$, where $\mathcal{E}, \mathcal{R}, \mathcal{A}, \mathcal{V}$ is the set of entities, relations, attributes and attribute values, respectively. 
The $\mathcal{T}^r = \{(h,r,t)\mid h,t \in \mathcal{E}, r \in \mathcal{R}\}$ is the set of relational triples.
The $\mathcal{T}^a = \{(e,a,v)\mid e\in \mathcal{E}, a\in \mathcal{A}, v\in \mathcal{V}\}$ is the set of attribute triples.

\textbf{Entity Alignment. }
Given the two KGs $\mathcal{G}_1$ and $\mathcal{G}_2$, the target of EA is to find a mapping between two KGs, i.e. $\mathcal{P} = \{(e, e') | e \in \mathcal{G}_1, e' \in \mathcal{G}_2\}$.
A set of alignment seeds $\mathcal{P}^s$ is used as training data.

\subsection{Overview}

In our TEA framework, we first transform an entity as textual sequences composed of its neighbors and attribute values, and then measure the similarity between a pair of cross-KG entities via a text entailment task on their sequences. Finally, we perform the entity alignment based on similarity. 

Now we elaborate on the textual entailment task. 
As shown in Figure \ref{fig: model_overview}, we first combine two sequences of cross-KG entities with a cloze-style template, and input the combined sequence into the PLM. Then, we tune the PLM with the entailment objectives to enlarge the positive entailment probability of the positive entity pairs.
The entailment probability $p(y|T(e,e'))$ is from one of the two proposed PLM-based entity aligners, NSP-Aligner or MLM-Aligner.

In practice, we find that the computationally cost is prohibitive to perform text entailment between all the entity pairs in two KGs. 
Therefore, besides the entailment objectives, we also tune the PLM simultaneously with the entity embedding-alignment objective, which minimizes the distance between the embeddings of the aligned entity pairs. 
For efficient EA inference, we first filter out the most similar candidates based on the embeddings learned from the embedding-alignment objective, and then re-rank these candidates via the entity similarity learned from the entailment objectives.

\subsection{Input construction}
\textbf{Sequence construction.}
We follow previous studies \citep{tang2021bertint,zhong2022sdea} to construct sequences with neighbors and attribute values, which contain rich semantics.
For entity $e$, the relational neighbors are $\mathcal{N}_e = \{ n | (e,r,n) \in \mathcal{T}^r\}$, and the attribute values are $\mathcal{V}_e = \{ v | (e,a,v) \in \mathcal{T}^a \}$.
We sort the $\mathcal{N}_e$ and $\mathcal{V}_e$ in alphabetical order by relation $r$ and attribute $a$ to form sequences respectively.
The sequences are denoted as 
$S^r(e)=\text{"}e,n_1,n_2,...,n_{|\mathcal{N}_e|}\texttt{[SEP]}\text{"}, n_i \in \mathcal{N}_e$ 
and $S^a(e)=\text{"}e,v_1,v_2,...,v_{|\mathcal{V}_e|}\texttt{[SEP]}\text{"}, v_i \in \mathcal{V}_e$.

\textbf{Entity-pair input.}
Existing PLM-based EA methods usually take the weighted hidden state of $\texttt{[CLS]}$ of single-entity input $x=\texttt{[CLS]}S(e)\texttt{[SEP]}$ for entity embedding. 
In our work, we propose to combine the sequences of two entities together and learn from their correlation.
The input could be denoted as $T(e,e') = \texttt{[CLS]}S(e)\texttt{[T]}S(e')$, where the $S(e)$ and $S(e')$ could be $S^r(e)$ or $S^a(e)$, and $\texttt{[T]}$ could be any templates.
We discuss the effect of templates in Section \ref{sec: effect_template}.

\textbf{Attention mask matrix.}
As shown in Figure \ref{fig: model_overview}, we design an attention mask matrix $M$  to implement the simultaneous tuning of the entailment objectives and the entity embedding-alignment objective, where the entailment mask $M_0$ exposes the whole entity-pair sequence to PLM and embedding masks $M_1$ and $M_2$ expose only one of the entities. 

\subsection{Training}
\textbf{Training set.}
In each epoch, we first construct a training set $\mathcal{D}=\{(e,e^+,e^-)|(e,e^+)\in \mathcal{P}^s, e^-\in \mathcal{G}_2, e^+ \ne e^-\}$, where each alignment seed $(e,e^+)$ from the training data $\mathcal{P}^s$ has a negative counterpart $e^-$.
Thus the model could be trained to distinguish the positive pair $(e,e^+)$ from the negative pair $(e, e^-)$.
We randomly select $e^-$ from the top entities in $\mathcal{G}_2$ with the highest embedding cosine similarity scores with $e$.
The embeddings for negative sample selection are obtained from the fixed PLM with single-entity input, and are consistent with the embeddings which are fine-tuned in the training phase with entity-pair input and embedding masks $M_1$ or $M_2$.

\textbf{Bi-directional training.}
For learning the bi-directional correlation between entities for alignment, we tune the PLM with the bi-directional sequences, i.e. $T(e, e')$ and $T(e', e)$.

\textbf{Cooperated training. }
For capturing the common correlation pattern of relational and attribute information, we tune the PLM with one epoch of relational input $T^r(e,e')$ and one epoch of attribute input $T^a(e,e')$ until convergence.

\subsection{Embedding-Alignment Objective}
The sequence $T(e,e')$ is tokenized and put into a pre-trained language model with the attention mask, such as multilingual BERT for cross-lingual EA.
We denote the obtained hidden states conditioned on the input sequence and attention mask $M_m$ as $H^m = \{\mathbf{h}_{\texttt{[CLS]}}^m,\mathbf{h}_1^m,...,\mathbf{h}_l^m,\mathbf{h}_{\texttt{[SEP]}}^m\} = \texttt{PLM}(T(e,e');M_m)$.

We obtain the embedding of entities following a standard fine-tuning paradigm. 
We obtain the hidden output of the PLM for the two entities $\mathbf{e} = \mathbf{W}_{\rm{emb}} \mathbf{h}^1_{\texttt{[CLS]}}$ and $\mathbf{e}' = \mathbf{W}_{\rm{emb}} \mathbf{h}^2_{\texttt{[CLS]}}$, where the $\mathbf{W}_{\rm{emb}} \in \mathbb{R}^{emb \times d}$ projects the hidden size of PLM $d$ to embedding size $emb$.
Then we apply the pairwise margin ranking loss in the embeddings of the training set as Equation \eqref{eq: loss_emb_margin} to minimize the distance between the positive entity pairs and maximize the distance of negative entity pairs.
The $d(\mathbf{e},\mathbf{e}')$ denotes the distance function between two entities and $m$ is a hyper-parameter that represents the margin between the positive and negative pairs. 
We use $l_2$ distance as distance function.
\begin{equation} \label{eq: loss_emb_margin}
\small
    \mathcal{L}_{mr} = \sum_{(e,e^+,e^-)\in\mathcal{D}} max \{ 0, d(\mathbf{e},\mathbf{e}^+) - d(\mathbf{e}, \mathbf{e}^-) + m \}.
\end{equation}

\subsection{Entailment Objectives}
For fully using the language modeling ability of PLMs, existing methods \citep{gao2021lmbff,sun2022nspbert} propose to model the downstream task as the pre-training tasks of PLM, i.e. NSP and MLM.
We propose two aligners based on the pre-training tasks of PLMs, i.e. NSP-Aligner and MLM-Aligner. 
Since we transform the EA task to a bi-directional text entailment task, we directly utilize NSP Head or MLM Head to represent if two entities entail each other, i.e. align to each other.
We denote the label space of entailment-style EA as $\mathcal{Y} = \{\texttt{align, not\_align}\}$.

\textbf{NSP-Aligner.} 
The origin NSP task predicts if the second sentence comes after the first sentence. 
For NSP-Aligner, the model predicts the probability of whether entity $e$ is after $e'$ and vice versa, to demonstrate the correlation of two entities. 
In this way, we can treat the entailment-style EA task as an NSP task.
As shown in Equation \eqref{eq: nsp_logit}, with the input of $T(e,e')$, the output of NSP head is the pre-softmax logit $p_{nsp}$, where $n \in \{\texttt{next, not\_next}\}$ respects to $\mathcal{Y}$, $\mathbf{W}_{\rm{nsp}} \in \mathbb{R}^{2 \times d}$ is the weight matrix learned by NSP task, and $\mathbf{h}^0_{\texttt{[CLS]}}$ is the hidden state of $\texttt{[CLS]}$ with the entailment mask $M_0$. 
\begin{equation} \label{eq: nsp_logit}
\small
\begin{aligned}
    p_{nsp}(y|T(e,e')) & = p(n | T(e,e')) \\
                 & = \mathbf{W}_{\rm{nsp}}(tanh(\mathbf{Wh}^0_{\texttt{[CLS]}}+\mathbf{b}))
\end{aligned}
\end{equation}

\textbf{MLM-Aligner. }
The origin MLM task predicts the masked token $\texttt{[MASK]}$ in the sequence.
For MLM-Aligner, the model learns a mapping from the label space to the set of individual words in the vocabulary, denoted as $\mathcal{M}: \mathcal{Y} \to \mathcal{V}$ with label word such as "\textit{Yes}" of "\textit{No}". 
In this way, we can treat the entailment-style EA task as an MLM task.
The MLM head fills the gaps $\texttt{[MASK]}$ with the label word probability as Equation \eqref{eq: mlm_logit}, where $\mathbf{W}_{\rm{mlm}} \in \mathbb{R}^{V \times d}$ projects the hidden state of PLM to the vocabulary size and $\mathbf{h}^0_{\texttt{[MASK]}}$ is the hidden state of $\texttt{[MASK]}$ with the entailment mask $M_0$.
\begin{equation} \label{eq: mlm_logit}
\small
\begin{aligned}
    p_{mlm}(y|T(e,e')) & = p(\texttt{[MASK]}=\mathcal{M}(y) | T(e,e')) \\
                 & = \mathbf{W}_{\rm{mlm}}\mathbf{h}^0_{\texttt{[MASK]}} + \mathbf{b}
\end{aligned}
\end{equation}

\textbf{Prompt bi-directional entailment loss. }In the training phase, we train the NSP-Aligner or MLM-Aligner with two losses. 
The first loss is a binary cross entropy loss for prompt entailment $\mathcal{L}_{pe}$ as shown in Equation \eqref{eq: loss_bi_entail} where $q(y|T(e,e')) = \texttt{softmax}(p(y|T(e,e')))$. 
We train the positive entity pair with positive label $1$ and the negative pair with negative label $0$.
We also add the reversed $\mathcal{L}'_{pe}$ with the input $T(e',e)$ for bi-directional modeling.
The final bi-directional entailment loss is $\mathcal{L}_{be} = \mathcal{L}_{pe} + \mathcal{L}'_{pe}$.
\begin{equation} \label{eq: loss_bi_entail}
\small
    \mathcal{L}_{pe} = \texttt{BCE}(q(y|T(e,e^+),1) + \texttt{BCE}(q(y|T(e,e^-)),0)
\end{equation}

\textbf{Prompt bi-directional margin loss. }
The second loss is the prompt margin ranking loss $\mathcal{L}_{pmr}$ as Equation \eqref{eq: loss_prompt_margin}, where the positive probability $p^{+}(y|T(e,e'))$ of positive entity pairs are enlarged compared to the negative pairs. 
The positive probability is $p_{nsp}^{+}(y|T(e,e')) = p(n=\texttt{next}|T(e,e'))$ for NSP-Aligner and $p_{mlm}^{+}(y|T(e,e')) = p(\texttt{[MASK]}=\textit{"Yes"}|T(e,e'))$ for MLM-Aligner.
We also use the bi-directional prompt margin loss as $\mathcal{L}_{bm} = \mathcal{L}_{pmr} + \mathcal{L}'_{pmr}$.
\begin{equation} \label{eq: loss_prompt_margin}
\small
\begin{aligned}
    \mathcal{L}_{pmr} =\sum_{(e,e^+,e^-) \in \mathcal{D}}&max\{0, p^{+}(y|T(e,e^-)) \\
        &- p^{+}(y|T(e,e^+)) + m \} \\
\end{aligned}
\end{equation}

The overall objective of TEA is the sum of three losses as Equation \eqref{eq: overall_loss}.
\begin{equation} \label{eq: overall_loss}
    \mathcal{L} = \mathcal{L}_{mr} + \mathcal{L}_{be} + \mathcal{L}_{bm}
\end{equation}

\subsection{Inference}
In the inference phase, we use entity embeddings for the first ranking. Then we use the PLM-based aligner, NSP-Aligner or MLM-Aligner, for re-ranking the hard samples with the candidates selected by the entity embeddings.

\textbf{Candidate entity selection.}
We use entity embeddings to select the candidate entity set. 
For each entity in $\mathcal{G}_1$, we retrieve the top fixed number of entities from $\mathcal{G}_2$ with the highest cosine similarity scores as candidate entity set $\mathcal{C}(e)$. The candidate number $|\mathcal{C}(e)|$ is hyper-parameter.

\textbf{Confidence-aware sample selection.}
We use the highest similarity score between $e$ and entities in $\mathcal{C}(e)$ as the embedding confidence score for sample $e$, denoted as $c(e) =max\{cos(\mathbf{e},\mathbf{e}')|e' \in \mathcal{C}(e)\}$. 
We assume that the test samples with lower confidence scores are harder samples for embeddings to obtain accurate results.
Then we re-rank the samples with lower confidence than a fixed threshold $c(e) < \delta$, with the positive probability $p^+(y|T(e,e'))$ of PLM-based aligner.
The samples with higher confidence use the similarity of embeddings as final alignment results.
The threshold $\delta$ is hyper-parameter.

\section{Experiments} \label{sec:experiments}

\begin{table*}[!t]
\centering
\resizebox{\textwidth}{!}{
\setlength{\tabcolsep}{1mm}{
\begin{tabular}{l|ccc|ccc|ccc|ccc|ccc}
\toprule
  {\multirow{2}{*}{Method}} & \multicolumn{3}{c}{$\rm{DBP_{ZH-EN}}$} & \multicolumn{3}{c}{$\rm{DBP_{JA-EN}}$} & \multicolumn{3}{c}{$\rm{DBP_{FR-EN}}$}& \multicolumn{3}{c}{$\rm{SRPRS_{EN-FR}}$} & \multicolumn{3}{c}{$\rm{SRPRS_{EN-DE}}$} \\
& H@1 & H@10 & MRR & H@1 & H@10 & MRR & H@1 & H@10 & MRR & H@1 & H@10 & MRR & H@1 & H@10 & MRR \\

\midrule
\multicolumn{16}{c}{\textit{Methods modeling relational triples and entity names}} \\
\midrule
RDGCN & 69.7 & 84.2 & 0.75 & 76.3 & 89.7 & 0.81 & 87.3 & 95.0 & 0.90 & 67.2 & 76.7 & 0.71 & 77.9 & 88.6 & 0.82 \\
HGCN  & 70.8 & 84.0 & 0.76 & 75.8 & 88.9 & 0.81 & 88.8 & 95.9 & 0.91 & 67.0 & 77.0 & 0.71 & 76.3 & 86.3 & 0.80 \\
CEA(Emb)   & 71.9 & 85.4 & 0.77 & 78.5 & 90.5 & 0.83 & 92.8 & 98.1 & 0.95 & 93.3 & 97.4 & 0.95 & 94.5 & 98.0 & 0.96 \\
CEA   & 78.7 & -   & -   & 86.3 & -   & -   & \textbf{97.2} & -   & -   & 96.2 & -   & -   & 97.1 & -   & -   \\
\midrule
FT-EA w/o $\mathcal{T}^a$   & 67.5 & 91.0 & 0.76 & 68.9 & 90.8 & 0.77 & 95.8 & 99.3 & \underline{\textbf{0.97}} & 96.7 & 98.8 & 0.97 & 97.0 & \underline{\textbf{99.1}} & \underline{\textbf{0.98}} \\
TEA-NSP w/o $\mathcal{T}^a$ & \underline{\textbf{81.5}} & \underline{\textbf{95.3}} & \underline{\textbf{0.87}} & \textbf{89.0} & \textbf{96.7} & \textbf{0.92} & \underline{\textbf{96.8}} & \textbf{99.5} & \textbf{0.98} & \underline{\textbf{97.3}} & \underline{\textbf{99.4}} & \underline{\textbf{0.98}} & \underline{\textbf{97.2}} & \textbf{99.6} & \underline{\textbf{0.98}} \\
TEA-MLM w/o $\mathcal{T}^a$& \textbf{83.1} & \textbf{95.7} & \textbf{0.88} & \underline{ \textbf{88.3}} & \underline{ \textbf{96.6}} & \underline{ \textbf{0.91}} & \underline{ \textbf{96.8}} & \underline{ \textbf{99.4}} & \textbf{0.98} & \textbf{98.1} & \textbf{99.5} & \textbf{0.99} & \textbf{98.3} & \textbf{99.6} & \textbf{0.99} \\

\midrule
\multicolumn{16}{c}{\textit{Methods modeling relational triples, attribute triples, and entity names}} \\
\midrule
AttrGNN & 79.6 & 92.9 & 0.85 & 78.3 & 92.1 & 0.83 & 91.9 & 97.8 & 0.91 & -   & -   & -   & -   & -   & -   \\
BERT-INT(name)& 81.4 & 83.5 & 0.82 & 80.6 & 83.5 & 0.82 & \textbf{98.7} & 99.2 & \textbf{0.99} & \underline{\textbf{97.1}} & 97.5 & \underline{\textbf{0.97}} & \underline{\textbf{98.6}} & 98.8 & \textbf{0.99} \\
SDEA  & 87.0 & 96.6 & 0.91 & 84.8 & 95.2 & 0.89 & 96.9 & 99.5 & \underline{0.98}  & 96.6 & 98.6 & \underline{\textbf{0.97}} & 96.8 & 98.9 & \underline{\textbf{0.98}} \\
\midrule
FT-EA & 85.4 & 95.7 & 0.89 & 83.2 & 93.4 & 0.87 & 95.7 & 99.0 & 0.97 & 96.4 & \underline{98.9}  & \underline{\textbf{0.97}} & 97.0 & 99.1 & \underline{\textbf{0.98}} \\
TEA-NSP & \textbf{94.1} & \textbf{98.3} & \textbf{0.96} & \textbf{94.1} & \textbf{97.9} & \textbf{0.96} & \underline{\textbf{97.9}} & \textbf{99.7} & \textbf{0.99} & \textbf{98.5} & \textbf{99.6} & \textbf{0.99} & \textbf{98.7} & \underline{\textbf{99.6}} & \textbf{0.99} \\
TEA-MLM & \underline{\textbf{93.5}} & \underline{\textbf{98.2}} & \underline{\textbf{0.95}} & \underline{\textbf{93.9}} & \underline{\textbf{97.8}} & \underline{\textbf{0.95}} & \textbf{98.7} & \underline{\textbf{99.6}} & \textbf{0.99} & \textbf{98.5} & \textbf{99.6} & \textbf{0.99} & \textbf{98.7} & \textbf{99.7} & \textbf{0.99}   \\
\bottomrule
\end{tabular}}}
\caption{Entity alignment performance on DBP15K and SRPRS. 
We highlight the \textbf{best} and the \textbf{\underline{second best}} results of each column. 
The "w/o $\mathcal{T}^a$" means training the model without modeling attribute information.
The TEA-NSP and TEA-MLM achieve the best or the second best in all metrics on all datasets.
}
\label{tab: main_results}
\end{table*}
\begin{table}[!b]
\centering
\resizebox{0.47\textwidth}{!}{
\begin{tabular}{cc|rrrrr|r}
\toprule
\multicolumn{2}{c}{Dataset}   & $|\mathcal{E}|$ & $|\mathcal{R}|$ & $|\mathcal{A}|$ & $|\mathcal{T}^r|$ & $|\mathcal{T}^a|$ & $|\mathcal{P}|$ \\
\midrule
\multirow{2}{*}{$\rm{DBP_{ZH-EN}}$} & ZH & 19,388 & 1,701 & 7,780 & 70,414 & 379,684 & \multirow{2}{*}{15,000} \\
& EN & 19,572 & 1,323 & 6,933 & 95,142 & 567,755 & \\
\multirow{2}{*}{$\rm{DBP_{JA-EN}}$} & JA & 19,814 & 1,299 & 5,681 & 77,214 & 354,619 & \multirow{2}{*}{15,000} \\
& EN & 19,780 & 1,153 & 5,850 & 93,484 & 497,230 & \\
\multirow{2}{*}{$\rm{DBP_{FR-EN}}$} & FR & 19,661 & 903 & 4,431 & 105,998 & 528,665 & \multirow{2}{*}{15,000}  \\
& EN & 19,993 & 1,208 & 6,161 & 115,722 & 576,543 \\
\midrule
\multirow{2}{*}{$\rm{SRPRS_{EN-FR}}$} & EN & 15,000 & 221 & 274 & 36,508 & 70,750 & \multirow{2}{*}{15,000} \\
& FR & 15,000 & 177 & 393 & 33,532 & 56,344 & \\

\multirow{2}{*}{$\rm{SRPRS_{EN-DE}}$} & EN & 15,000 & 222 & 275 & 38,363 & 62,715 & \multirow{2}{*}{15,000} \\
& DE & 15,000 & 120 & 185 & 37,377 & 142,506 &\\

\bottomrule
\end{tabular}
}
\caption{Datasets statistics for EA.}
\label{tab: datasets}
\end{table}

\subsection{Experimental Settings}
\textbf{Datasets.}
To evaluate the proposed method, we conduct experiments on two widely used EA datasets: DBP15K \citep{sun2017jape} and  SRPRS \citep{guo2019RSN}. 
\textbf{DBP15K} is the most commonly used EA dataset and consists of three cross-lingual EA subsets, which are Chinese-English (ZH-EN), Japanese-English (JA-EN), and French-English (FR-EN).
\textbf{SRPRS} is a sparse EA dataset with much fewer triples and consists of two cross-lingual EA subsets, which are English-French (EN-FR) and English-German (EN-DE). 
The dataset statistics of DBP15K and SRPRS are listed in Table \ref{tab: datasets}.
Consistent with previous studies, we randomly choose 30\% of the samples for training and 70\% for testing.

\textbf{Evaluation metrics.}
We use Hits@K (K=1,10), which is the accuracy in top K predictions, and Mean Reciprocal Rank (MRR), which is the average reciprocal ranking of ground-truth entity, as evaluation metrics.
The higher Hits@K and higher MRR indicate better performance.

\textbf{Implementation details.}
We implement our approach with Pytorch and Transformers \citep{wolf2020huggingface}.
We use BERT \citep{devlin2019bert} as the PLM for cross-lingual EA following \citet{liu2020attrgnn,tang2021bertint,zhong2022sdea}.
The information for evaluating TEA is one of the relational and attribute information which performs higher Hits@1 in the validation set, i.e. attribute for DBP15K and relational for SRPRS. 
The training is early stopped after 3 epochs of no improvements of Hits@1 in the validation set.
We conduct the experiments in Ubuntu 18.04.5 with a single NVIDIA A6000 GPU with 48GB of RAM.

\textbf{Baselines.}
To comprehensively evaluate our method TEA, the baselines are grouped into two categories according to the input information.
Since we construct the sequences with entity names, we mainly compare TEA with the method that also models entity names.
(1) The methods modeling relational triples and entity names: RDGCN \citep{wu2019RDGCN}, HGCN \citep{wu2019HGCN}, CEA \citep{zeng2020CEA}. 
(2) The methods modeling relational triples, attribute triples, and entity names: AttrGNN \citep{liu2020attrgnn}, BERT-INT \citep{tang2021bertint}, SDEA \citep{zhong2022sdea}. 
For BERT-INT which uses entity descriptions, we replace the descriptions with entity names for a fair comparison following \citet{zhong2022sdea}.

We construct a baseline FT-EA, which learns and inferences with the entity embeddings for alignment results.
FT-EA could be seen as TEA w/o textual entailment objectives and re-ranking.
We report the results of TEA with two PLM-based aligners, TEA-NSP and TEA-MLM. 
We also ablated the attribute sequence (w/o $\mathcal{T}^a$) to compare with the baselines of group (1).

\subsection{Comparison with Baselines.} \label{sec: compare_baselines}
We compare our method with the baselines and the results are presented in Table \ref{tab: main_results}.

\textbf{Comparison with group (1).}
Compared with methods modeling relational triples and entity names, TEA-NSP and TEA-MLM achieve the best or the second best in all metrics on all datasets. 
Even on $\rm{DBP_{ZH-EN}}$ where baselines fail to perform well, TEA-MLM outperforms the baselines by at most 4.4\% in Hits@1 and 11\% in MRR. 
Moreover, compared with FT-EA, the re-ranking with NSP-Aligner and MLM-Aligner brings significant improvements, at most 20.1\% in Hits@1 and 15\% in MRR improvements.

The TEA-NSP and TEA-MLM perform comparably on DBP15K and TEA-MLM performs better than TEA-NSP on SRPRS.
The reason could be that MLM-Aligner is more competitive in the low-resource setting \citep{gao2021lmbff} since the SRPRS dataset has fewer triples. 
We will look into EA under the low-resource setting in the future.

\textbf{Comparison with group (2).}
Compared with methods modeling heterogeneous triples and entity names, TEA performs the best or the second best in all metrics.
The TEA-NSP outperforms the baselines by 9.3\% in Hits@1 and 7\% in MRR at most, and outperforms the FT-EA by 10.9\% in Hits@1 and 9\% in MRR at most.
We could observe that BERT-INT(name) \citep{tang2021bertint} performs the best or the second best in some metrics on the FR-EN, EN-FR, and EN-DE alignment. 
The reason could be that BERT-INT relies more on the similarity between entity names, and English shares many similar expressions with French and German.
Thus BERT-INT's performance declines on the alignment between less-alike languages.

\begin{table}[!t]
\centering
\resizebox{0.85\linewidth}{!}{
\setlength{\tabcolsep}{1mm}{
\begin{tabular}{l|ccc}
\toprule
\multirow{2}{*}{} & \multicolumn{3}{c}{$\rm{DBP_{ZH-EN}}$} \\
&  Hits@1  & Hits@10  & MRR  \\
\midrule
TEA-NSP &  94.1 & 98.3  & 0.96 \\
\midrule
TEA-NSP w/o $\texttt{[T]}$  & 92.6 & 97.7  & 0.95 \\
TEA-NSP w/o $\mathcal{L}_{be}$   & 90.3 & 97.4  & 0.93 \\
TEA-NSP w/o $\mathcal{L}_{bm}$   & 93.2 & 98.0  & 0.95 \\
TEA-NSP w/o $\mathcal{T}^r$  & 90.1 & 97.1  & 0.93 \\
MLM-FT-EA  & 85.2 & 95.2  & 0.89 \\
\bottomrule
\end{tabular}}}
\caption{Ablation study on $\rm{DBP_{ZH-EN}}$.
The $\texttt{[T]}$ means templates. $\mathcal{L}_{be}$ and $\mathcal{L}_{bm}$ means the prompt bi-directional entailment loss and margin loss.
$\mathcal{T}^r$ means relational information.
MLM-FT-EA is a variation of FT-EA where the entity embeddings are obtained in MLM-style.
}
\label{tab: ablation}
\end{table}

TEA on SRPRS in group (1) and (2) are both evaluated with relational sequences.
With extra attribute information, TEA in group (2) outperforms the TEA w/o $\mathcal{T}^a$ in group (1).
It demonstrates that by modeling the common correlation pattern of the heterogeneous information with the PLM-based aligners, the extra attribute information would enhance the alignment of relational information.
On the contrary, without the modeling of the common correlation, the performance of FT-EA slightly declines or stays the same on the SRPRS dataset than FT-EA w/o $\mathcal{T}^a$.

The TEA-NSP are comparable but slightly better than TEA-MLM in group (2). 
The reason could be that the interaction modeling of two aligners is similar, but NSP-Aligner is better with sentence-pair input than MLM-Aligner since NSP is designed to process sentence pairs.

\begin{table}[!t]
\centering
\resizebox{\linewidth}{!}{
\setlength{\tabcolsep}{0.2mm}{
\begin{tabular}{l|ccc|ccc}
\toprule
\multirow{2}{*}{Template $T(e,e')$}  & \multicolumn{3}{c|}{TEA-NSP} & \multicolumn{3}{c}{TEA-MLM} \\
 & H@1  & H@10  & MRR   & H@1  & H@10  & MRR   \\
\midrule
\multicolumn{7}{c}{Hard templates}\\
\midrule
$S(e)$\pattern{? \texttt{[MASK]}.}$S(e')$ & 93.3 & 98.1 & 0.95 & 93.2 & 98.1 & 0.95 \\

$S(e)$\pattern{? \texttt{[MASK]}. I know that} $S(e')$ & 93.6 & 97.8 & 0.95 & 93.4 & 97.8 & 0.95 \\

$S(e)$\pattern{? \texttt{[MASK]}. I think that} $S(e')$ & 92.3 & 97.4 & 0.94 & 93.2 & 97.8 & 0.95 \\

\midrule
\multicolumn{7}{c}{Soft templates}       \\
\midrule
$S(e)$\pattern{\texttt{[MASK][P$_{0}$]}...\texttt{[P$_{l}$]}}$S(e')$, l=1 & 94.1 & 98.3 & 0.96 & 92.8 & 97.8 & 0.95 \\

$S(e)$\pattern{\texttt{[MASK][P$_{0}$]}...\texttt{[P$_{l}$]}}$S(e')$, l=2 & 93.4 & 97.8 & 0.95 & 93.2 & 97.9 & 0.95 \\

$S(e)$\pattern{\texttt{[MASK][P$_{0}$]}...\texttt{[P$_{l}$]}}$S(e')$, l=3 & 92.8 & 97.8 & 0.95 & 93.3 & 98.2 & 0.95 \\

$S(e)$\pattern{\texttt{[MASK][P$_{0}$]}...\texttt{[P$_{l}$]}}$S(e')$, l=4 & 92.5 & 97.8 & 0.95 & 93.5 & 98.2 & 0.95        \\
\bottomrule
\end{tabular}}}
\caption{Effect of templates on $\rm{DBP_{ZH-EN}}$. For hard templates, we manually design some templates. For soft templates, we use the special token $\texttt{P}_l$ following \citet{ding2021promptfget}, where $l$ is a hyper-parameter. }
\label{tab: templates}
\end{table}

\subsection{Ablation Study}
We conduct the ablation study as shown in Table \ref{tab: ablation}.

\textit{Q1: Is the cloze-style template necessary for NSP-Aligner? } 
Since most prompt-learning methods use the cloze-style templates to form an MLM task rather than an NSP task, thus we remove the cloze-style template in the NSP-Aligner with TEA-NSP w/o $\texttt{[T]}$, i.e. only use $\texttt{[SEP]}$ token to divide the sequences of two entities.
The performance declines 1.5\% in Hits@1 compared to the TEA-NSP, which shows that the template could also enhance the performance of NSP-Aligner.

\textit{Q2: Are the entailment objectives necessary? }
The ablation of two entailment losses $\mathcal{L}_{be}$ and $\mathcal{L}_{bm}$ results in a decrease of 3.8\% and 0.9\%, respectively. 
Thus two losses both enhance the re-ranking performance and the binary cross-entropy loss enhances more than the margin loss.

\textit{Q3: Do the relational sequences and attribute sequences enhance each other? }
The TEA-NSP and the TEA-NSP w/o $\mathcal{T}^r$ are both evaluated by attribute information.
By modeling the extra relational information, the performance of evaluating with attribute information increases by 4.0\% in Hits@1, which means the modeling of relational information enhances the modeling of the attribute information.
Moreover, the analysis in Section \ref{sec: compare_baselines} shows the reversed enhancement.
They demonstrate that by modeling the common correlation of relational and attribute information in a unified manner would enable mutual enhancement.

\textit{Q4: Is the interaction of entity-pair necessary? } 
We construct MLM-FT-EA, a variation of FT-EA, to ablate the entity-pair interaction with reservation of the prompt learning.
Inspired by recent progress in sentence embedding \citep{jiang2022promptbert}, we use a cloze-style template \pattern{This sentence of “$S(e)$” means \texttt{[MASK]}.} to obtain entity embeddings with MLM-FT-EA.
The performance of MLM-FT-EA is similar to FT-EA.
It shows that the entity-pair interaction is the most important component in TEA rather than the prompt-learning paradigm.

\begin{figure}[!t]
    \centering
    \includegraphics[width=0.49\textwidth]{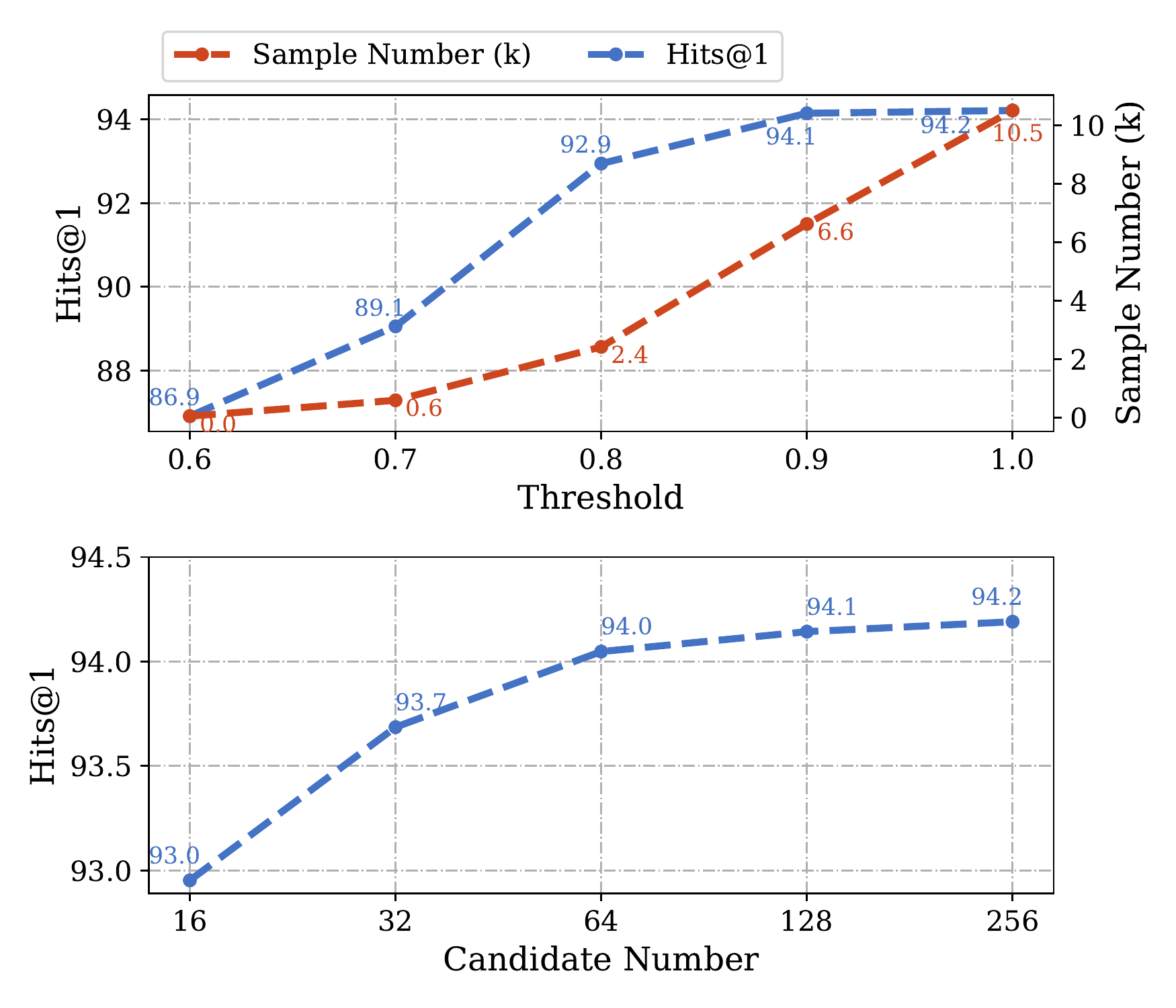}
    \caption{Re-ranking parameter analysis conducted by TEA-NSP on $\rm{DBP_{ZH-EN}}$.}
    \label{fig: rerank_paras}
\end{figure}

\subsection{Effect of Templates} \label{sec: effect_template}
In this section, we study the effect of templates in TEA.
As stated by previous studies \citep{gao2021lmbff,tam2021pattern}, the templates have impacts on the performance of prompt-learning oriented tasks. 
We design both hard templates and soft templates on $\rm{DBP_{ZH-EN}}$ dataset.
The hard templates are manually designed, while the soft templates have a varying number of learnable special prompt tokens following \citet{ding2021promptfget}.
As shown in Table \ref{tab: templates}, the templates could affect the performance of EA considerably. 
For hard templates, the \pattern{I know that} improves the performance the most.
For soft templates, the TEA-NSP needs fewer special tokens  while the TEA-MLM needs more.

\subsection{Effect of Re-ranking Parameters} \label{sec: effect_reranking}
Figure \ref{fig: rerank_paras} shows the hyper-parameter analysis of the re-ranking process of TEA. 
The sample number is the number of entities in $\mathcal{G}_1$ to be re-ranked by the PLM-based aligners.
With a higher threshold, more samples are re-ranked and the performance of EA is better.
When threshold $\delta = 0.9$, the re-ranking samples are $37\%$ less than re-ranking all the samples ($\delta = 1.0$)  but the performance is similar and the re-ranking time cost are highly reduced.

The candidate number is the number of entities in $\mathcal{G}_2$ that are most likely to be the ground truth. 
With more candidates, the performance is better.
The reason could that the ground truth entity is more likely to be in the candidate set when the candidate set is larger.
Moreover, even with only 16 candidates, the performance of TEA in Hits@1 exceeds the FT-EA by $7.6\%$.

\begin{table}[t!]
\small
\centering
\resizebox{0.49\textwidth}{!}{
\begin{tabular}[t]{p{4cm}p{4cm}}
\toprule
\textbf{Before Re-ranking}     & \textbf{After Re-ranking}
\\
\midrule
\multicolumn{2}{l}{\textbf{\textit{Visualization of Attention Weights}}} \\
\multicolumn{2}{c}{
\raisebox{-0.95\totalheight}{
    \includegraphics[width=1.1\linewidth]{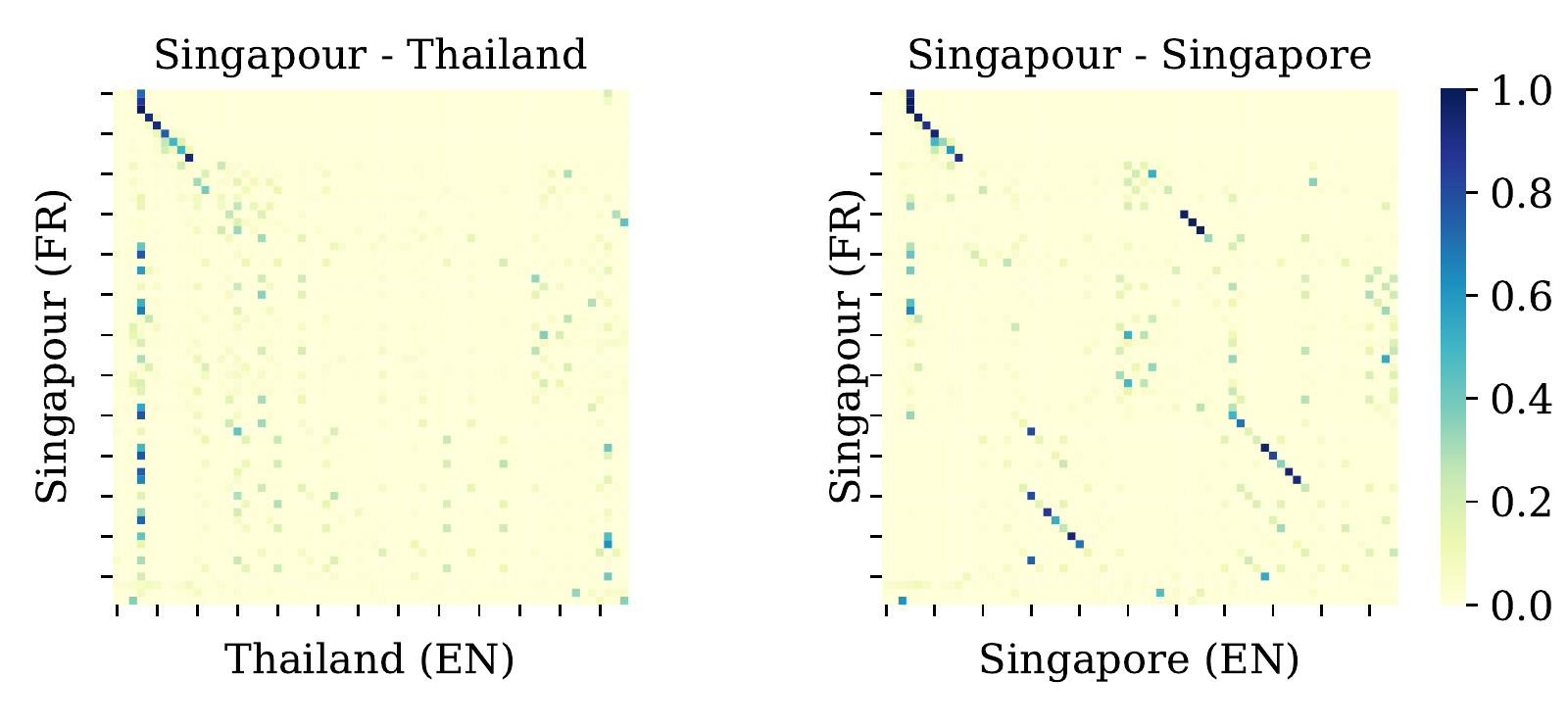} 
}
}
\\
\midrule
\multicolumn{2}{l}{\textbf{\textit{Rankings of Entity Candidates}}} \\
\entity{Thailand}, Malaysia, Hong Kong, Khmer language, Khuang Aphaiwong, Cambodia, Patna, \entity{Singapore}, ...
& 
\entity{Singapore}, Indonesian language, Javanese language, West Sumatra, Malaysia, \entity{Thailand}, Jakarta Cambodia, Patna, ... \\
\bottomrule
\end{tabular}
}
\caption{The attention weights of entity pair with highest alignment probability and the entity rankings before and after re-ranking.
}
\label{tab: case}
\end{table}

\subsection{Case Study}
We conduct a case study as shown in Table \ref{tab: case}, trying to find the aligned entity of \textit{Singapour (FR)}.
The entity ranking conducted by embeddings shows the best-aligned entity is \textit{Thailand (EN)}.
However, by re-ranking the candidates with PLM-based aligners, the fine-grained interaction between entities is explicitly modeled.
As shown in the visualization, the \textit{Singapour (FR)-Singapore (EN)} pair has more attentive sub-sequences (darker diagonal short lines) while the unaligned pair \textit{Singapour (FR)-Thailand (EN)} have not. 
Moreover, the aligned entity is ranked first place by the PLM-based aligner.

\section{Conclusion}
To address the limitations of the existing EA method, the lack of interaction and heterogeneous embedding spaces, 
we propose a unified textual entailment framework for entity alignment called TEA.
We transform the origin relational triples and attribute triples of an entity into textual sequences and model the EA task as a bi-directional textual entailment task between the sequences of cross-KG entities. 
We propose two kinds of PLM-based aligners to capture the fine-grained correlation between entities with two kinds of sequences in a unified manner.
The entailment probability is used for measuring entity similarity and ranking the entity candidates.
Experiment results on five cross-lingual datasets show that TEA outperforms existing EA methods and enables the mutual enhancement between the heterogeneous information.

\section*{Limitations}

Despite that TEA achieves some gains for EA, TEA still has the following limitations:

First, TEA has a higher computation cost than the embedding-based EA methods in the re-ranking phase, since TEA process entity-pair input for modeling the interaction between them.
For reducing time costs, we adopt the confidence-aware re-ranking strategy to reduce the number of re-ranking samples and candidates.
However, the inference time cost is still higher than the embedding-based methods.
In addition, the candidate selection may be limited in some corner cases if the ground truth entity is not ranked in the top $|\mathcal{C}|$ similar entities calculated by entity embeddings.
We will further explore efficient approaches which could cover the corner cases.

Second, the alignment of relational information of TEA requires the entity names to construct sequences. 
However, the entity names are not always available in some EA datasets, such as the Wikidata KG in OpenEA Benchmark \citep{sun2020benchmarking}.
In that case, TEA can use the attribute sequences without entity names for entity alignment.
Though TEA w/o $\mathcal{T}^r$ can achieve competitive performance as shown in Table \ref{tab: ablation}, it still limits the application of TEA.
We will further explore PLM-based approaches to align the relational information without the requirement of entity names.

\section*{Acknowledgements}
This research is supported by the Natural Science Foundation of Tianjin, China (No. 22JCJQJC00150, 22JCQNJC01580),
the National Natural Science Foundation of China
(No. 62272250, U1936206, 62002178), 
Tianjin Research Innovation Project for Postgraduate Students (No. 2022SKYZ232), 
and the Fundamental Research Funds for the Central Universities (No. 63232114).

\bibliography{mybib,custom,anthology}
\bibliographystyle{acl_natbib}

\end{document}